\documentclass[letterpaper, 10 pt, conference]{ieeeconf}  % Comment this line out if you need a4paper

\IEEEoverridecommandlockouts                              % This command is only needed if 
\usepackage{colortbl}
\usepackage{hyperref}

\usepackage[colorinlistoftodos,prependcaption,textsize=small]{todonotes}

\usepackage{amsmath}
\usepackage{textcomp}
\usepackage{amssymb}
\usepackage{mathtools}
\usepackage{bm}

\title{\LARGE \bf
Quasi-Direct Drive for Low-Cost Compliant Robotic Manipulation
}

\author{David V. Gealy$^{1}$, Stephen McKinley$^{2}$, Brent Yi$^{3}$, Philipp Wu$^{1,3}$, \\
Phillip R. Downey$^{1}$, Greg Balke$^{3}$, Allan Zhao$^{3}$, Menglong Guo$^{1}$,  \\ 
Rachel Thomasson$^{1}$, Anthony Sinclair$^{1}$, Peter Cuellar$^{1}$, Zoe McCarthy$^{3}$,  and Pieter Abbeel$^{3}$%
\thanks{\hspace{-8pt}Affiliations and Corresponding Authors:}
\thanks{$^{1}$Mechanical Engineering, University of California, Berkeley
        {\tt\small dgealy@berkeley.edu}}%
\thanks{$^{2}$Industrial Engineering \& Operations Research, University of California, Berkeley
        {\tt\small mckinley@berkeley.edu}}%
\thanks{$^{3}$Electrical Engineering \& Computer Science, University of California, Berkeley
        {\tt\small abbeel@berkeley.edu}}%  
}

\begin{document}

\maketitle
\thispagestyle{empty}
\pagestyle{empty}

%%%%%%%%%%%%%%%%%%%%%%%%%%%%%%%%%%%%%%%%%%%%%%%%%%%%%%%%%%%%%%%%%%%%%%%%%%%%%%%%
\begin{abstract}
Robots must cost less and be force-controlled to enable widespread, safe deployment in unconstrained human environments.
We propose Quasi-Direct Drive actuation as a capable paradigm for robotic force-controlled manipulation in human environments at low-cost.
Our prototype - \textit{Blue} - is a human scale 7 Degree of Freedom arm with 2kg payload. \textit{Blue} can cost less than \$5000.
We show that \textit{Blue} has dynamic properties that meet or exceed the needs of human operators: the robot has a nominal position-control bandwidth of 7.5Hz and repeatability within 4mm.
We demonstrate a Virtual Reality based interface that can be used as a method for telepresence and collecting robot training demonstrations. Manufacturability, scaling, and potential use-cases for the \textit{Blue} system are also addressed.
Videos and additional information can be found online at \href{https://berkeleyopenarms.github.io/}{berkeleyopenarms.github.io}.
\end{abstract}
%
%%%%%%%%%%%%%%%%%%%%%%%%%%%%%%%%%%%%%%%%%%%%%%%%%%%%%%%%%%%%%%%%%%%%%%%%%%%%%%%

\section{Introduction}
\label{sec:intro}

\subsection{Problem Definition and User Needs}

The future of robotic manipulation is in unconstrained environments such as warehouses, homes, hospitals, and urban landscapes. These robots must operate with dexterity and safety alongside people despite imperfect actuation, lapses in sensing, and unmodeled contacts.
Unlike traditional position-controlled manipulators, force-controlled robots can robustly react to unpredicted interactions without incurring damage to the environment or the robot itself. 
We believe the current class of compliant manipulators are too expensive or lack sufficient performance to complete useful tasks in human environments. 
% Adoption of force-controlled robots has been limited due to cost, but as others have noted ``increased affordability can lead  to wider adoption, which in turn can lead to faster progress"  \cite{quigley2011low}.  
This paper presents a fully realized paradigm for a low-cost Quasi-Direct Drive (QDD) manipulator and discusses considerations taken for the design and manufacturing of this system.

\begin{figure}[t]
\centering
% \vspace{8pt}
\includegraphics[width=1\linewidth]{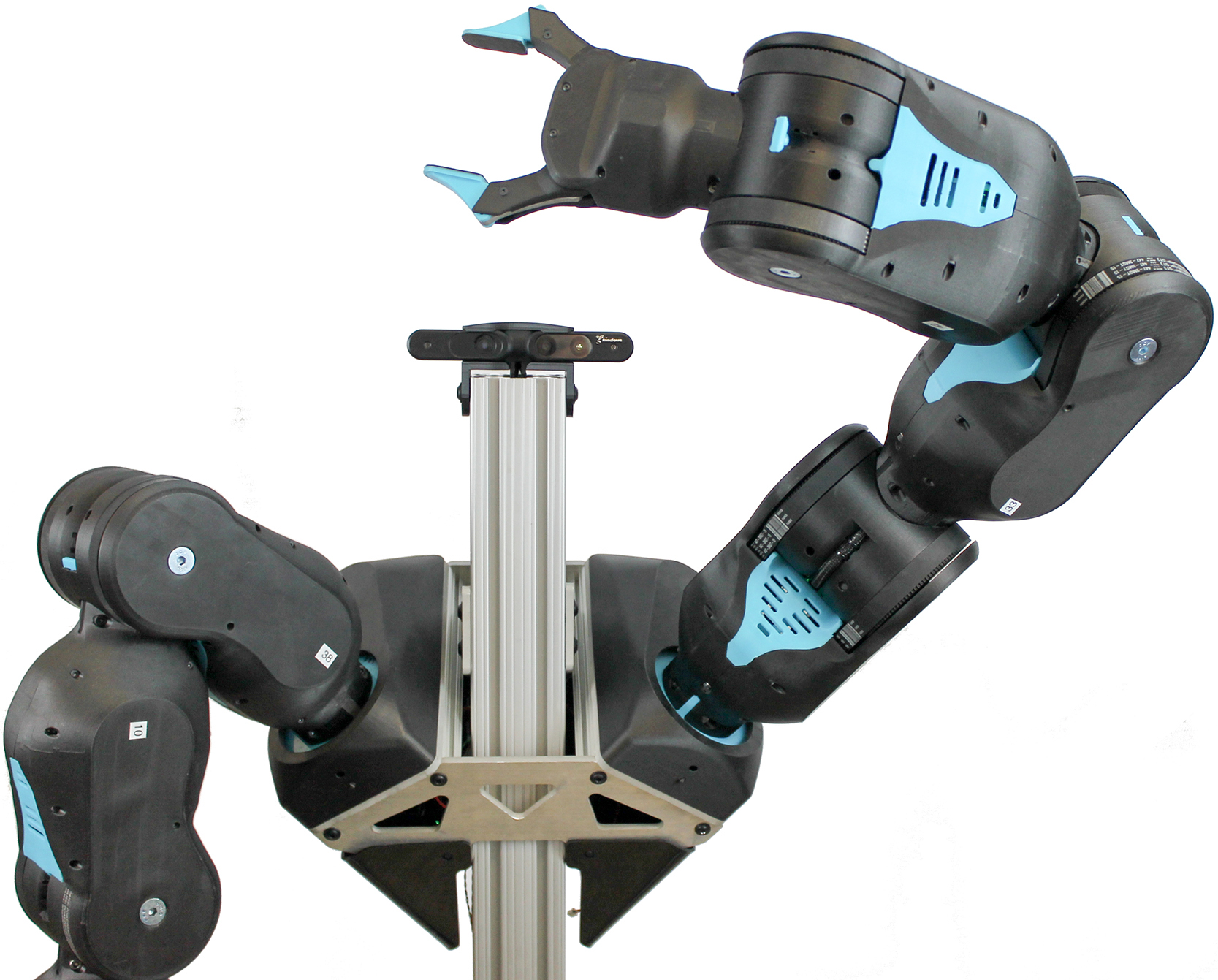}
\vspace{-6pt}
\caption{Unconstrained automation using Quasi-Direct Drive actuation.
The \textit{Blue} manipulator is a 7 Degree-of-Freedom robotic arm that is human-sized, compliant, has a 2kg payload, and can cost less than \$5000 per arm to end-users at scale.
\textit{Blue} is designed for unconstrained environments and for interactions with humans. Position repeatability is within 4mm and bandwidth exceeds human level. \bf{[video on website]}}
\label{fig:intro}
\vspace{-16pt}
\end{figure}

% % The success of classic robotic manipulation in traditional automation applications relies on high mechanical bandwidth and repeatability in concert with highly structured environments to rapidly perform useful tasks. 
Our design goals %considerations %stem from observing 
%aim to 
support recent trends in AI-based control methods. We believe these control methods can be more widely applied in human environments if force-controlled robots are made affordable.

% % Significant inspiration comes from the suggestion that tasks completed through human teleoperation can be learned and repeated \cite{yu2018one}. % this doesn't reference teleoperation 
% Learning from Demonstration (LfD) algorithms use human operators to provide demonstrations of tasks \cite{schulman2016learning}. These demonstrations can be acquired through VR teleoperation \cite{Koganti:2018:VRU:3170427.3186500} \cite{zhang2017deep}. 
% A kinematically-anthropomorphic robot (7 Degree of Freedom comprising 3 in the shoulder, 1 in the elbow, and 3 in the wrist) as shown in Figure \ref{fig:intro}, can mimic human motions and better enable LfD in human environments.

% %Another promising avenue for learned robot control is 
% Reinforcement Learning (RL) is an iterative method which seeks to maximize a given reward through the repeated refinement of a learned policy.
% If robot costs decrease, policy refinement can be accelerated efficiently by having multiple systems iterate in parallel \cite{levine2018learning}.

A kinematically-anthropomorphic robot (7 Degree of Freedom comprising 3 in the shoulder, 1 in the elbow, and 3 in the wrist) as shown in Figure \ref{fig:intro}, can better mimic human motions, allowing better maneuverability in human environments, and enabling more intuitive teleoperation. This can be useful in Learning from Demonstration (LfD), where human operators provide demonstrations of tasks through methods including Virtual Reality (VR) teleoperation \cite{Koganti:2018:VRU:3170427.3186500} \cite{zhang2017deep}.

If robot cost is reduced, iterative methods such as Reinforcement Learning (RL) which seek to maximize a given reward through the repeated refinement of a learned policy can be accelerated efficiently by allowing multiple systems to run policy refinement and iterate in parallel \cite{levine2018learning}.

Requirements for high robot repeatability may be less important for both teleoperation and learning based methods that use visual feedback \cite{finn2016deep} \cite{duan2016benchmarking}.
%Requirements for high robot repeatability may be less important for both RL-based manipulation that uses visual feedback \cite{finn2016deep} \cite{duan2016benchmarking}, and LfD that relies on expert demonstrations through teleoperation. \
Additionally, recent work utilizing domain randomization for AI-based policy generation suggests that lower-precision hardware can be used for grasping tasks \cite{mahler2017dex}.
%Requirements for robot repeatability may not be as important for RL-based manipulation that uses visual feedback \cite{finn2016deep} \cite{duan2016benchmarking}.
%
Training for both LfD and RL often leads to collisions between the robot and environment \cite{deisenroth2011learning}. 
Compliant robots can mitigate damage by controlling interaction forces. 

These considerations shift the focus of designing hardware away from the constraints of highly structured manufacturing environments (which depend on robots with high repeatability and high bandwidth), and instead moves toward the broader question:

\textbf{What hardware paradigms will most enable useful automation in unconstrained real-world human environments at low cost?}
\\ 
\\
\noindent \textit{\textbf{ Contributions}}: In this paper we present:
\begin{itemize}%[noitemsep, leftmargin=*, label=\arabic*.]
    \item our design criteria for useful robotic manipulation in unconstrained environments,
    \item an implementation of a robot arm that satisfies the above set of specifications, 
    \item evaluation of the physical characteristics of our new design, \item work towards DFM (design for manufacturing), and
    \item production cost analysis.
\end{itemize}

%%%%%%%%%%%%%%%%%%%%%%
\subsection{Defining a Useful Robotic Manipulator}
\label{subsection:definitions}

We define a design paradigm that enables \underline{useful}\textsuperscript{a}, \underline{low-cost}\textsuperscript{b} robotic arms capable of \underline{manipulation tasks}\textsuperscript{c} in unconstrained environments.

a) We define \underline{useful} in metrics similar to humans:
human-size, 7 Degrees of Freedom , 2kg payload, safe, compliant, and with a repeatability under 10mm.

b) We define \underline{low-cost} as: pricing below \$$5000$ to an end user for a manufacturing run of more than $1500$ arms.

c) A partial set of \underline{tasks} to consider includes: unloading a dishwasher, stocking a refrigerator, floor decluttering, opening doors, opening microwave ovens, sorting packages, physical stroke rehabilitation, folding laundry, cleaning windows, bed making, and bathroom cleaning. We demonstrate the robot in kitchen cleaning, table decluttering, telepresence, and machine tending.

\subsection{Defining Useful Bandwidth and Payload}
Super-human bandwidth and payload capabilities enable high speed and high precision in constrained industrial automation tasks. 
However, if the goal is to safely manipulate household objects through human teleoperation while minimizing cost, performance trade-offs have to be made. This motivates seeking new definitions for \underline{useful} bandwidth and payload metrics for our design. 

\begin{figure}[]
\centering
\vspace{8pt}
\includegraphics[width=1\linewidth]{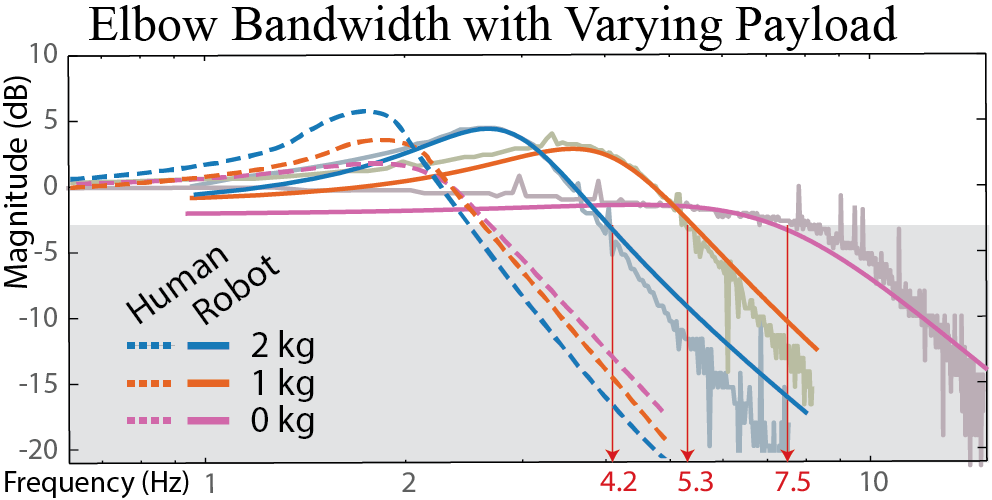}
\vspace{-9pt}
\caption{Our robot, \textit{Blue}, is designed for human-like motion. Position bandwidth of the elbow joint is compared to a similar test of the biceps brachii in humans \cite{aaron1976comparison}. Intersections between curves and the grey region represent frequencies beyond effective control (-3dB) for that loading condition. Raw data is shown as faded. Curves shown in solid were fit using a second order transfer function with a constant time delay.}
\label{fig:elbowBodePlot}
\vspace{-10pt}
\end{figure}

Bandwidth is a measure of an actuator's ability to deliver force (or control position) at higher frequencies. We believe a manipulator designed for human teleoperation can be perceived as useful as long as the robot's effective bandwidth is greater than that of the (human) user. As a lower bound for this design: studies on human muscle (biceps brachii) characteristics show that maximum effective position bandwidth is 2.3 Hz as found by Aaron et al \cite{aaron1976comparison}. See Figure \ref{fig:elbowBodePlot} for a comparison of human bandwidth characteristics and \textit{Blue's} properties.

Rated payloads for commercially available robots often cover conservative loading conditions: guaranteeing high-bandwidth trajectory tracking in worst-case positions under continuous operation while holding a maximum `rated' payload. 
We instead define a \underline{useful} payload as one that can cover the largest set of outlined \underline{tasks}, at human speeds. For example, position bandwidth can suffer under high payload, since dexterity at high payload is not universally required for the tasks considered in Section \ref{subsection:definitions}.

Drawing inspiration from nature, we consider how humans subconsciously minimize energy output during manipulation tasks \cite{mi2009optimization}.
Human arms have 1:1 arm mass to payload ratio (about $4$kg:$4$kg), 
but humans cannot maintain full payload constantly (100\% duty cycle). 
For object manipulation, humans have poor steady-state (RMS) force output, yet have high `burst power' capability. 
Extending this to robot design, overall robot mass and inertia can be reduced if maximum loads are assumed as peaks of short-duration effort rather than requirements for continuous operation.
In Section \ref{sec:performance}.E and Figures \ref{fig:heatPowerTotal},\ref{fig:heatPowerEach} we describe considerations for robots operating within a thermally limited paradigm.

\subsection{Examining Low-Cost Design Constraints}

Our goal is to lower the cost of general purpose robotic manipulators to the point where we can place a robot on the desk of every researcher in our group (the Robot Learning Lab at UC Berkeley). For this to be possible, we believe a system would have to be approximately equal in cost to a high performance research computer ($<\$5000$).

\section{Related Work}
\label{sec:relWork}

\subsection{Compliance in Robotic Systems}
Compliance is the ability for a robot to exhibit \textit{low impedance}: moving when disturbed by an outside force. %A system that is compliant is sometimes referred to as having \textit{low impedance}. 
A rigid non-compliant (high impedance) robot can be dangerous to operate near humans and destructive to itself or its environment during collisions.
However, an entirely compliant robot will not be able to deftly manipulate objects nor respond to high frequency commands (low bandwidth).

Compliance can be passively inherent in systems or actively added to otherwise non-compliant systems.
\textit{Active compliance} can be achieved through sensing of output torques and feedback control and is found in series elastic actuation (SEA) \cite{pratt1997stiffness} and modern `cobots' \cite{albu2007dlr}.
\textit{Passive compliance} is a characteristic of systems that can be driven by external forces with no use of feedback control.

Passive compliance can be achieved with backdrivable transmissions, wherein external forces applied at the output act on the motor and can be 'sensed' by measuring motor currents.
Backdrivability enables highly robust torque control % with a minimum number of sensors and low-cost components 
because the motor also acts as the torque sensor. Co-locating the sensor and actuator significantly eases dynamic stability problems present in force control \cite{eppinger1992three}.

High bandwidth actuation combined with inherently backdrivable transmissions allows a robot controller to select impedance (high or low) \cite{seok2012actuator}, helping match the unpredictable needs of real-world environments \cite{mason1981compliance}.
In the scope of our work, passive compliance is inherent within backdrivable actuation.

\subsection{Force-Controlled Manipulators at Human Payload}

\subsubsection{Industry Solutions}
Kuka's LBR has excellent closed-loop strain-based force control \cite{bischoff2010kuka} and sells for upwards of \$$67000$. The similar Franka Emika arm is available for \$$29900$. Rethink Robotic's `Baxter' was \$$25000$ (for two arms) and has been replaced by a single 7-DOF arm called `Sawyer' available for \$$29000$. Currently all robots mentioned above except Baxter use harmonic drives which can be made backdrivable with additional sensors but are not inherently compliant.

\subsubsection{Backdrivable Research Solutions}
The Barrett WAM (\$$135000$) is a highly-backdrivable manipulator accomplishing useful payload by placing all actuators in the base and using low-friction cable transmissions \cite{townsend1993mechanical}. The Willow Garage PR2 (\$$400,000$ for 2 armed fully integrated mobile manipulator) achieved backdrivability through a gravity compensation mechanism, allowing the undersizing of actuators \cite{wyrobek2008towards}. The high cost of these platforms was influenced by complex design approaches.

\subsection{Existing Low-cost Manipulators}
Quigley's Low-Cost Manipulator is an example of robot design built for manipulation research which makes careful design trade-offs balancing elements such as cost, compliance, and payload \cite{quigley2011low}. Although Bill of Materials (BoM) part cost is estimated at (\$$4135$), manufacturing costs and complexity were not accounted for.

Other low-cost manipulators (with or without compliance) are currently achieved through reduced Degrees of Freedom \cite{yamamoto2018human} \cite{eaton2016c}, and/or the use of off-the-shelf hobby servos and have significantly reduced payload \cite{7Bot_2015}.

% We explore a novel paradigm for low-cost compliant manipulation by relaxing constraints on robot open-loop-precision and control-bandwidth.
% This approach is reinforced by successful
\subsection{Actuation Schemes}

Successful implementations of series elastic robots have shown that useful tasks can still be completed despite lower mechanical and control bandwidths \cite{edsinger2004domo}.
However, it is not clear that existing SEA actuator solutions can be made low-cost. Rethink Robotics Baxter is the closest realization of this paradigm at $25000$ for a two-arm system.

% Passive compliance can be achieved with naturally backdrivable transmissions. 
% Backdrivability enables highly robust torque control with a minimum number of sensors and low-cost components because the motor also acts as the torque sensor. Co-locating the torque sensor and actuator significantly eases dynamic stability problems present in force control \cite{eppinger1992three}.
Backdrivable actuation holds promise for robots in unconstrained environments and enables selectable impedance with robust force control. 
Direct-drive is the most backdrivable, but high motor masses in the arm make high-DoF systems impractical.
Recently, Quasi-Direct Drive (QDD) actuators (transmission ratios $<$ 1:10) have been used for legged locomotion and have the desirable properties of low friction, high backdrivability, toughness, simplicity, robust force control and selectable impedance \cite{kalouche2016design}. The primary drawback of this actuation method is reduced torque-density \cite{seok2012actuator}. 

\section{Design for Low-Cost Compliant Manipulation}
\label{sec:design}
A full characterization of our design is shown in Table \ref{tableProperties}.

\begin{table}[t]
\vspace{4pt}
\caption{Physical properties of the \textit{Blue} manipulator}
\label{tableProperties}
\vspace{-9pt}
\centering
\includegraphics[width=1\linewidth]{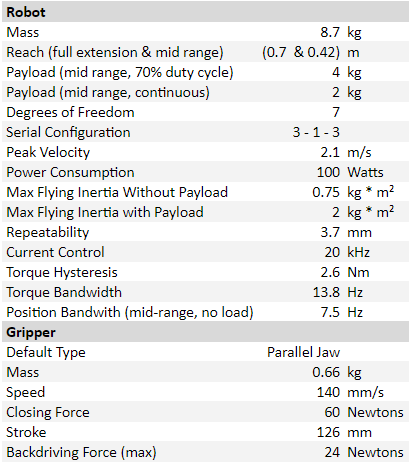}
\end{table}

\begin{figure}[]
\centering
\vspace{-6pt}
\includegraphics[width=1\linewidth]{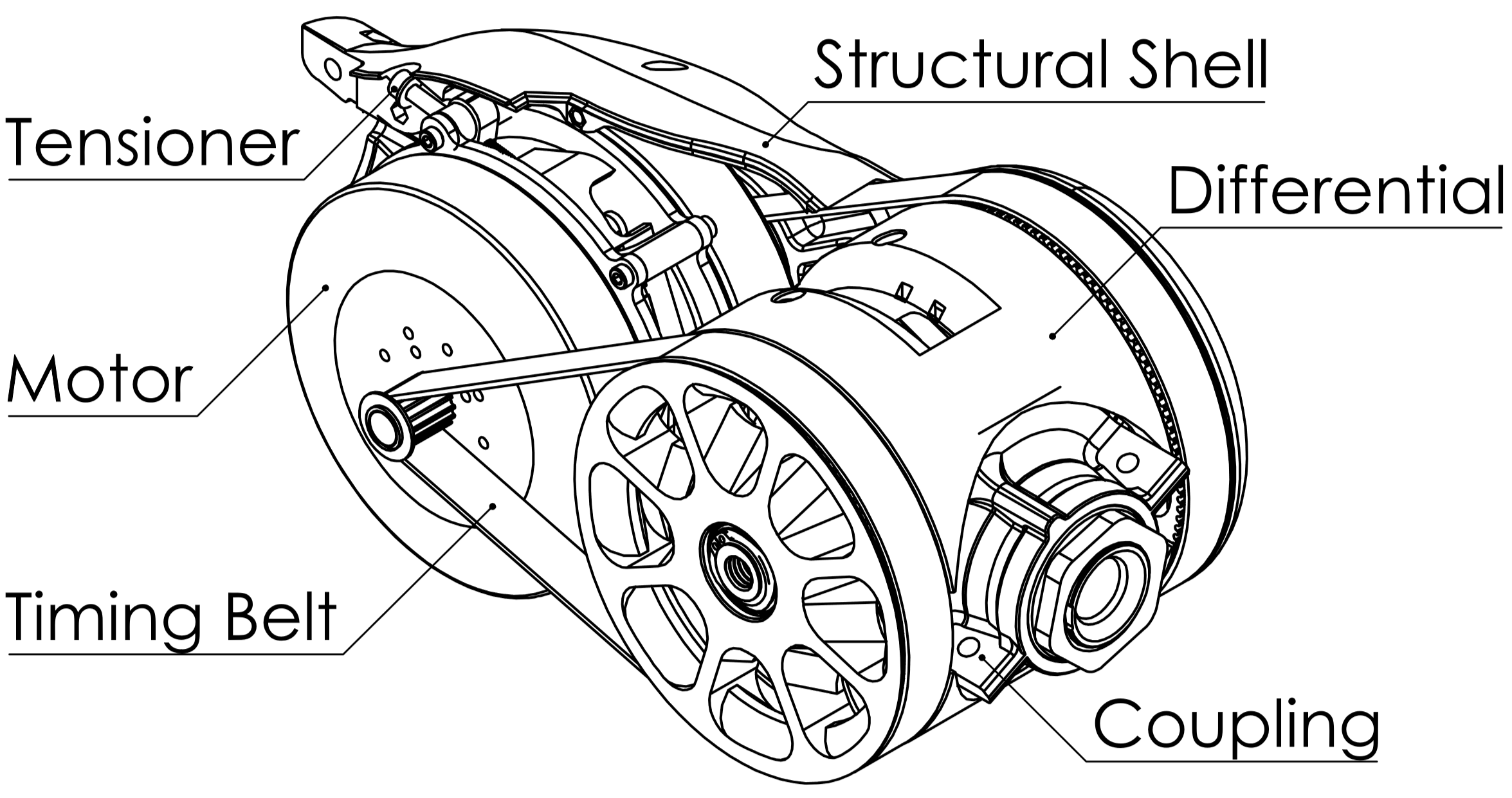}
\vspace{-9pt}
\caption{Internal view of a single 2-DoF geared differential module.}
\label{fig:cutawayView}
\vspace{-10pt}
\end{figure}

\subsection{Quasi-Direct Drive Actuation}

QDD was chosen for the \textit{Blue} system because it can achieve backdrivability in a wide range of transmission options (Gears, Belts, Cables, etc.), and has adequate torque density. 
Large gap-radius brushless outrunner gimbal motors from iFlight (see Table \ref{prototypeCost}) were selected for their exceptional Km density at relatively low cost. Outrunners have higher torque density at the cost of reduced thermal dissipation and increased inertia \cite{sensinger2011exterior}. Thermal considerations are considered in Section \ref{sec:performance}.

\subsection{Differential Timing Belt Transmissions}
Timing belt transmissions were chosen over cables because of their relative ease of assembly and tensioning, durability, allowance for continuous rotation, efficiency ($>$95\%), low backlash, and high backdrivability. 15mm wide GT3 belts with fiberglass tension elements were chosen with 3mm pitch to maximize the feasible single-stage gear ratio, transmitting power from a 16 tooth pinion to a 114 tooth output pulley resulting in a 7.125:1 single-stage reduction.
Each link of the robot has a 2-DoF differential output, combining two planar QDD timing belt transmissions into output pitch and roll motions as seen in Figure \ref{fig:cutawayView}. 
% Coupled differential actuation has been used in many robotic systems ranging from zero-backlash cable differentials of the inline variety \cite{townsend1993mechanical} and the offset variety \cite{edsinger_2016patent}, to geared differentials \cite{wyrobek2008towards}. %\cite{scorbotDatasheet} \cite{edsinger2004domo} \cite{bornich2016halodi} \cite{metta2008icub}
Benefits of differential drive include partial load sharing when splitting induced gravitational loads.
%as shown in Figure \ref{fig:differentialSharing}. 

An advantage of timing belts is their ability to transmit power over distance. Shifting the motor mass towards the shoulder reduces gravity induced torques and flying inertia, helping mitigate poor torque density inherent to QDD actuation. 
A conservative comparison is produced by locating a summed motor and transmission mass at each DoF, then calculating flying inertia and gravity induced torques about the shoulder. 
Shifting the motors back using timing belts results in an approximate 30\% reduction in both gravity induced torque about the shoulder, and 30\% reduction in flying inertia.

%We initially designed \textit{Blue} with cable differentials to avoid potential backlash. 
%However, g

Geared differentials were chosen for their simplicity, reliability, impact resistance, continuous rotation, and lower part count. Because the differentials operate at low speed, large plastic teeth can be used under preloads with success. 
% Benchtop tests indicate maximum cycle count in excess of 200,000. While \textit{Blue}'s open-loop force control is minimally affected by backlash, mechanical impedance would be adversely affected.

\subsection{Modular Structural Shells}

\textit{Blue} is a robot designed for potential interaction with humans. The modular, repeated structural shells are designed to house, protect, and support all components of the arm while concentrating design complexity into a few injection-moldable parts that handle structure, safety from pinch points, and mechanical coupling between stages. The base of each shell is a two-bolt clamp that resists torque in all directions. The coupling between shells is limited in diameter to avoid potential finger-pinch points. 

\begin{figure}[h]
\centering
\vspace{-6pt}
\includegraphics[width=1\linewidth]{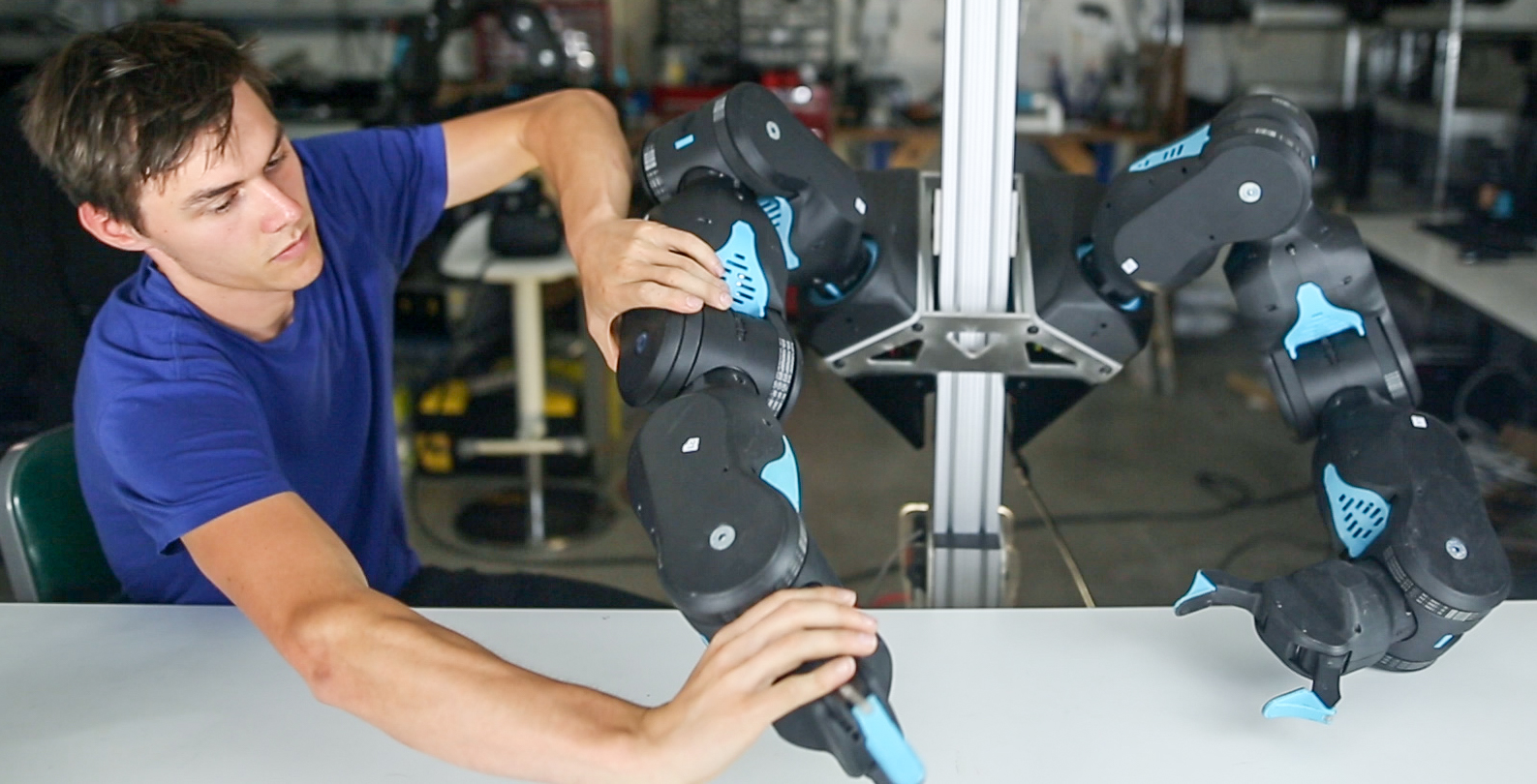}
\vspace{-12pt}
\caption{The \textit{Blue} arm is designed to minimize pinch points. Safety as a constraint during design heavily impacts the final form of a robot that will interact with human environments. \bf{[video on website]}}
\label{fig:pinchPoints}
\vspace{-10pt}
\end{figure}

Belt tension is applied by pivoting the servos assemblies away from the output pulleys using a lead screw. A single tension point balances loads on both timing belts. A drawback to this approach is that passive heat transfer from motor to environment is throttled to RMS 10 Watts per stage because the motors are isolated from thermal conductors. This limitation is surmounted using a fan in the base.

\subsection{Base and Gripper}
A 1-DoF timing belt base was developed to create a 3-DoF shoulder. While differential load sharing is removed, mounting the servomotor to a large aluminum base greatly increases both thermal mass and heat dissipation. 

A low-cost parallel jaw gripper was designed and implemented as shown in Figure \ref{fig:pinchPoints}. Comparable end-effectors used in research are often $>$\$5,000 USD and would defeat the purpose of a low-cost paradigm. 
A servo module (same as in the rest of the arm) drives a backdrivable lead screw that actuates the four-bar-linkage fingers through a rack-and-pinion.
Despite an increasing diversity of gripper paradigms, we chose parallel jaws for their predictability, robustness, simplicity (low cost), and ease of simulation \cite{mahler2017dex}.

\subsection{Motor Drivers and Sensors}
\textit{Blue's} Quasi-Direct Drive actuation utilizes a single driver board per servo with all sensors co-located to minimize wiring complexity, connector failure points, and manufacturing cost.
Custom motor drivers were developed for \textit{Blue} \cite{zhao2018design}.
Each driver board is equipped with the following sensors:
14-bit absolute on-axis magnetic encoding for motor commutation and robot position sensing;
12-bit current sensing for closed-loop current (and thus torque) control of each servomotor;
a 3-axis accelerometer for state estimation, collision detection and control as well as start-up robot calibration as envisioned in \cite{quigley2010low}, and
temperature sensors for thermal monitoring and shutdown if needed.

\begin{figure}[]
\centering
\vspace{8pt}
\includegraphics[width=1\linewidth]{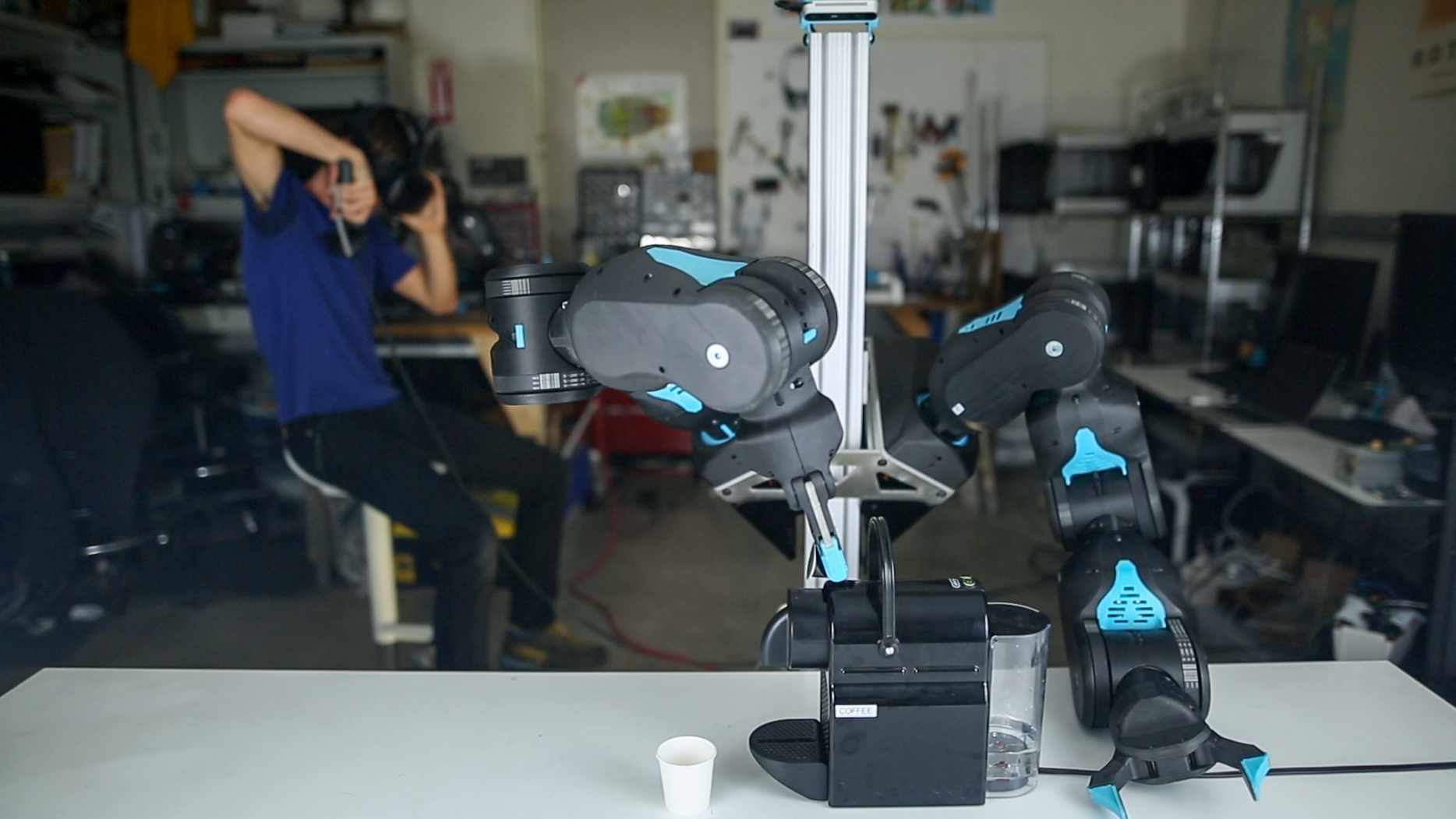}
\vspace{-9pt}
\caption{\textit{Blue} completely teleoperates an espresso machine. A Virtual Reality operator (in background) pilots \textit{Blue} using an HTC Vive system through a Unity bridge. A predictable 7-DoF `elbows out' configuration is interpreted from 6-DoF Vive controller pose. Visual feedback is provided by an Intel RealSense D415 depth camera. \bf{[video on website]}}
\label{fig:vrTeleop}
\vspace{-10pt}
\end{figure}

\section{System Integration}
\label{sec:integration}

% \subsection{Power Supply}
Power is supplied from a two-quadrant 48V 8A MeanWell switching regulator. Peak power can be anticipated at 250 W instantaneous, and 25 W continuous with no payload. A custom reverse current shunt circuit protects the low-cost power supply from reverse current-flow.

\subsection{Control and Communication}
\textit{Blue's} control system is built around a central control computer (currently an Intel NUC, BOXNUC7I3BNK) that runs Ubuntu Linux and makes heavy use of the \texttt{ros\_control} \cite{chitta2017ros_control} framework. The computer determines actuator torque commands and sends them to each servomotor through a shared RS485 bus running at 170Hz with a 1Mbps data rate. Firmware updates, driver configuration, and motor calibration also leverage the same RS485 bus.

PID joint control is computed with feed-forward gravity and Coriolis compensation torques. 
Each servomotor locally runs real-time current control at 20kHz. 
%Additional control modes include local PD controllers which help maximize peak impedance without introducing stability issues caused by two-way communication latency.

% \subsection{Communication Protocol}
% A custom 1Mbps communication protocol over RS485 was developed to connect all motor drivers drivers in parallel. With the current packet size, control frequency is 160Hz.
% A shared RS485 bus is used for all communication with our motor drivers.
% At each cycle of the robot control loop, the software builds a single packet that contains all motor commands and a request for actuator state information. This packet is multicasted to the motor boards, each of which responds with a complete summary of its current state and sensor readings. 
% The resulting control frequency is 170Hz with 8 boards and a 1Mbps data rate.

% To run the standard stack, we can use a raspberry pi for control. Once we bring networking into the loop the system falls apart. However, we showed that we could record and replay motions over the Pi which is more than enough to say that it is computationally capable to run the arm.

%%%%%%%%%%%%%%%%%%%%% P.Wu Math %%%%%%%%%%%%%%%%%%%%%%%%
%%%%%%%%%%%%%%%%%%%%%%% START %%%%%%%%%%%%%%%%%%%%%%%%%%

\subsection{Inverse Kinematics}
%Having a end effector controller that smoothly manipulates the arm in a natural manner is of critical importance for human teleoperation.
% This renders commonly used trajectory planning control schemes like RRT's less ideal. 

% The use of human teleopertion using VR requires a custom inverse kinematics solution that allows for realtime control and continuous joint state manipulation.

% Option 1
Controlling the 7-DoF end effector in real time through a 6-DoF VR teleoperation interface (as demonstrated in Figure \ref{fig:vrTeleop}) requires a computationally efficient algorithm with joint state continuity.
% Option 2
% The use of human teleopertion using VR requires a computationally efficient inverse kinematics solution that allows for realtime control and continuous joint state manipulation.
An iterative inverse kinematics solver is used with a secondary joint state objective to constrain the arm's redundant degree of freedom \cite{oussama_sicilian_robot_handbook}. 
Teleoperation was made more intuitive by setting the secondary objective to match a human's resting position, with elbows naturally oriented. The joint error to this pose is optimized in the null space of the 7-DoF manipulator Jacobian.

Telepresence allows the user to interact with others remotely and in real time. Operation of machinery (Figure \ref{fig:vrTeleop}) is possible using the VR interface and \textit{Blue's} compliance allows it to be safely manipulated in human environments.

% From work done with the PR2 \cite{schulman2016learning}, it has been shown that learning by demonstration is quite possible and highly applicable with compliant arms.

\section{Performance Metrics and Experiments}
\label{sec:performance}

\begin{figure}[t]
\centering
\vspace{4pt}
\includegraphics[width=1\linewidth]{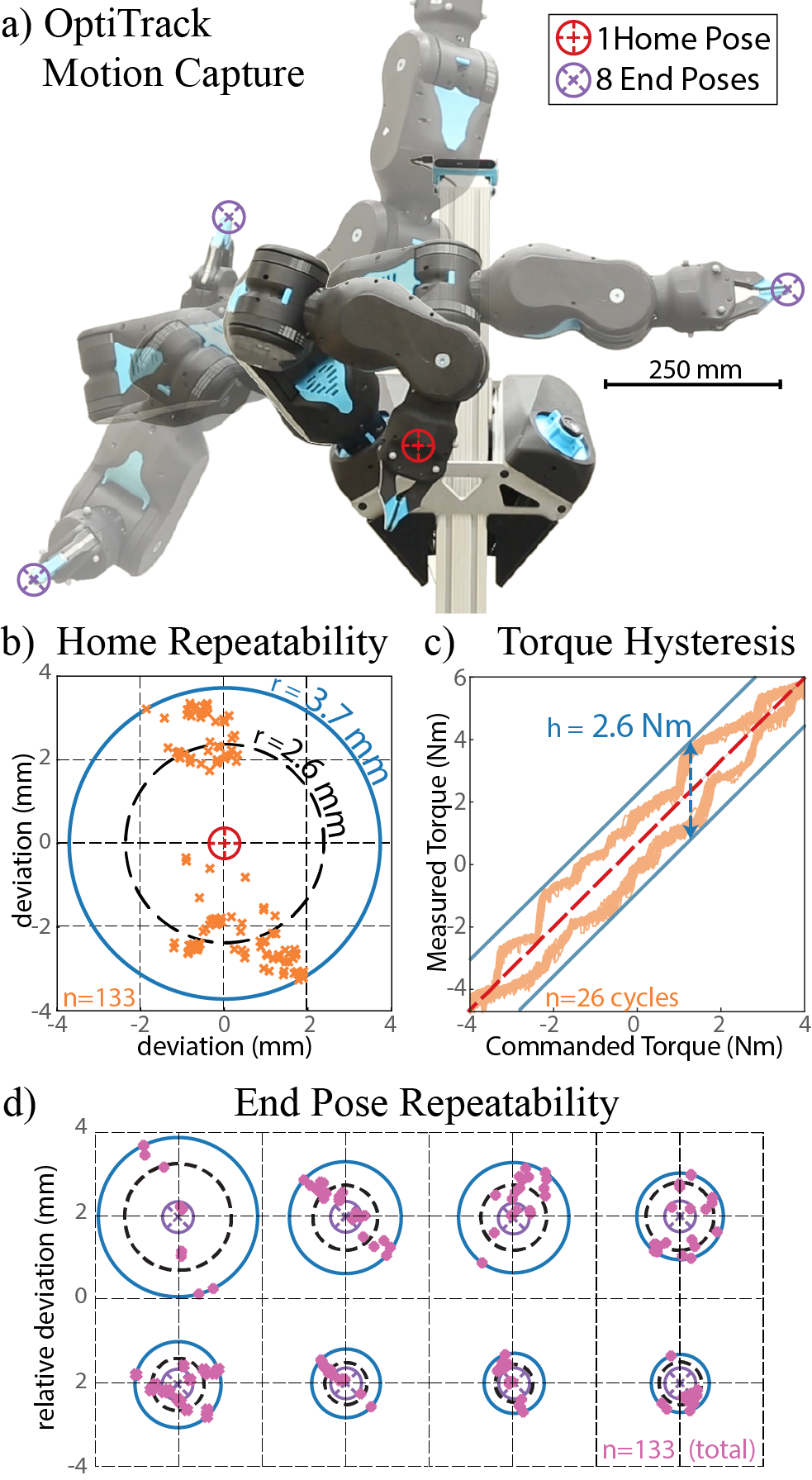}
% \vspace{-9pt}
\caption{%Repeatability Tests: 
\textbf{a)} %OptiTrack IR reflective markers were placed on robot end effector. 
The arm was commanded to one of 8 End Poses (purple targets) before returning to Home Pose (red target). Position is recorded after return-to-home.
\textbf{b)} Home-pose repeatability is shown in the plane of highest variance. %All of $133$ motions returned within a circle of radius $3.7$ mm, the average deviation was $2.6$ mm.
\textbf{c)} Torque hysteresis caused by friction encourages a bimodal distribution in return-to-home points dependant on direction.
\textbf{d)} End-pose repeatability shown in the plane of highest variance for each set. Scale matches inset b, poses are arranged for clarity. Poses closer to the robot's torso tended to have higher variance. \bf{[video on website]}}
\label{fig:repeatability}
\vspace{-10pt}
\end{figure}

\subsection{Repeatability Experiments}
Position-control repeatability was measured (similarly to \cite{quigley2011low}) by moving from a `home' position to one of eight predefined dwell locations in the robot workspace (chosen in random order) and then back to home as shown in Figure \ref{fig:repeatability}.a. Motion was recorded by an OptiTrack motion Capture system at 100Hz. The standard deviation of the home pose was (0.89, 2.2, 1.6) mm  for the (x, y, z) axis and (0.53, 0.29, 0.09) degrees for the (roll, pitch, yaw) axis. 
Figure \ref{fig:repeatability}.b shows the distribution of home dwell points during $133$ motions sliced across the plane of highest variance. All trials fell within a radius of 3.7 mm and the average deviation from the center of home poses was 2.6 mm. Figure \ref{fig:repeatability}.d shows the distribution of end effector poses around each end point. Higher repeatability can be achieved through adding output joint encoders or visual feedback.

\subsection{Static Torque Hysteresis}
% Friction (present in all actuators) limits backdrivability, preventing commanded torque from perfectly matching output torque. This results in a hysteresis band, the height of which represents a bound on the static, non-desirable stochastic mapping of motor and output torques.

Friction (present in all actuators) limits backdrivability and degrades the potential for a motor to act as a torque sensor.
%and adds stochasticity to what would ideally be a perfect one-to-one bidirectional mapping between motor and output torques. 
This friction results in a torque hysteresis band, the height of which represents a bound on the uncertainty of the mapping between commanded torque and actual output torque.  %Designing for backdrivability involves minimizing this band while still allowing for adequate torque density. 
The static torque hysteresis band was measured by locking the actuator output, slowly cycling motor torque, and measuring output torque. The results of this show a worst case bound of 2.6 Nm for a full 2-DoF differential actuating an output roll as shown in Figure \ref{fig:repeatability}.c. 

Motor cogging torque accounts for 0.47 Nm of a total measured 0.89 Nm backdriving torque per single belt transmission. Torque ripple caused by motor cogging can be seen in \ref{fig:repeatability}.c. Although cogging torque is currently lumped with friction, methods exist to reduce this effect through additional calibration and feed-forward control \cite{piccoli2016anticogging}.

\begin{figure}[]
\centering
\vspace{8pt}
\includegraphics[width=1\linewidth]{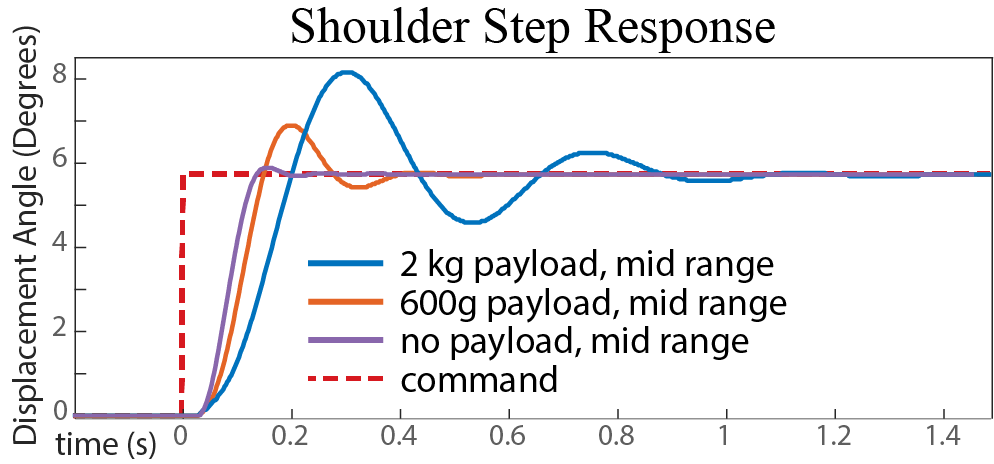}
\vspace{-9pt}
\caption{Actuator response to an instantaneous change in position command. A 6 degree change in position (red dashed line) was requested under three loading conditions. This overshoot correspond to 0.15, 1.1, and 2.3 degrees respectively. \bf{[video on website]}}
\label{fig:stepResponse}
\vspace{-10pt}
\end{figure}
% a smaller area within the shape indicates lower losses
% because friction losses are directional 
% the maximum height across the hysteresis band indicates the `deadband' of the actuator.

% Hysteresis tests measure the difference between torque requested of the actuator and torque delivered by the actuator while slowly cycling continuously through a range of torques.

\subsection{Position-Control Experiments}
% Regardless of a future-proof vision, position control performance remains of utmost importance for robots to achieve utility now. Robot repeatability, position control bandwidth, and 
A test cell was constructed to secure one 2-DOF modular link and measure output position in real time while incorporating stiffness of the belt transmission (1.3 kNm/rad), and lumping of differential, 3D-printed plastic shells, and 3D-printed joint coupling stiffness (lumped at 1.2 kNm/radian). 
Masses were held vertically to avoid directional transmission pre-load from gravity and a rotary encoder was used to record arm translation via cable capstan transmission.
%and small angle approximation. 
%cable capstan transmission was used to record arm translation as rotation of a rotary encoder via small angle approximation.

\subsubsection{Step Responses}

As seen in Figure \ref{fig:stepResponse}, step responses for the shoulder lift were performed with inertias of (0.13, 0.27, 0.76) kgm$^2$ representing (0, 0.6, 2) kg payloads at mid-range (50-70\% robot reach). For a 6 degree step command, overshoot is (0.15, 1.1, 2.3) degrees, resulting in (0.1, 0.8, 2.8) cm  of end-effector overshoot at mid-range. Underdamped dynamics can be partially handled through smoother trajectories, provided by either teleoperation or trajectory optimization.

\subsubsection{Position Bandwidth}

\begin{figure}[]
\centering
\vspace{8pt}
\includegraphics[width=1\linewidth]{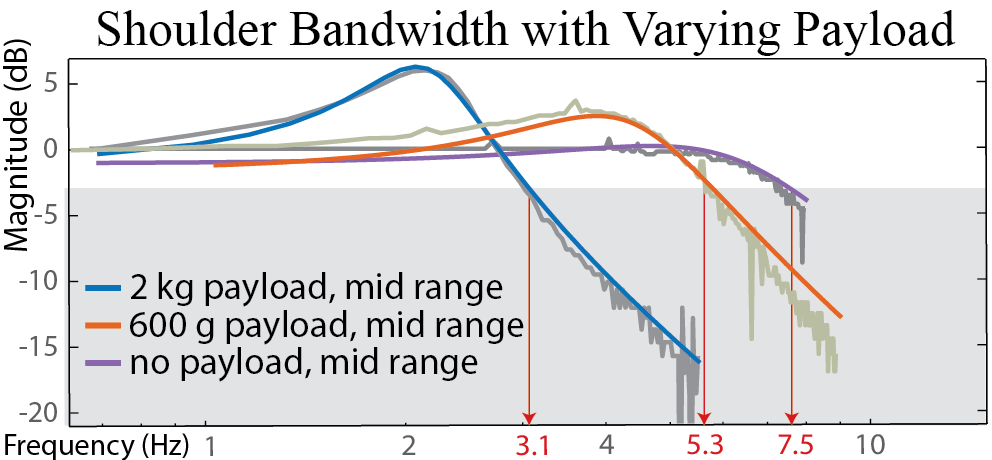}
\vspace{-9pt}
\caption{Shoulder bandwidth with payload and arm fully extended is evaluated as a worst-case measure of the system's ability to respond to high frequency commands while loaded. Intersections between curves and the grey region represent frequencies beyond effective control (-3dB) for each loading condition. Raw data is shown as faded, curves shown in solid were fit using a second order transfer function with a constant time delay. \bf{[video on website]}}
\label{fig:shoulderBodePlot}
\vspace{-10pt}
\end{figure}

Position control bandwidth describes the maximum frequency with which an actuator can effectively track a pose command. Figure \ref{fig:elbowBodePlot} compares robot position bandwidth to that of humans' biceps brachii with varying payloads, while Figure \ref{fig:shoulderBodePlot} describes position-control bandwidths for the shoulder for various mid-range loads. 

\subsection{Torque Bandwidth Experiments}
Torque bandwidth is a measure of how quickly commanded torque can propagate through a transmission, resulting in a change in output torque. High torque bandwidth coupled with backdrivable transmission enables selectable impedance control, wherein the manipulator dynamics can be rapidly changed to best fit the interacting environment. Torque bandwidth was measured to better understand actuator performance in human environments.

\begin{figure}[h]
\centering
\vspace{8pt}
\includegraphics[width=1\linewidth]{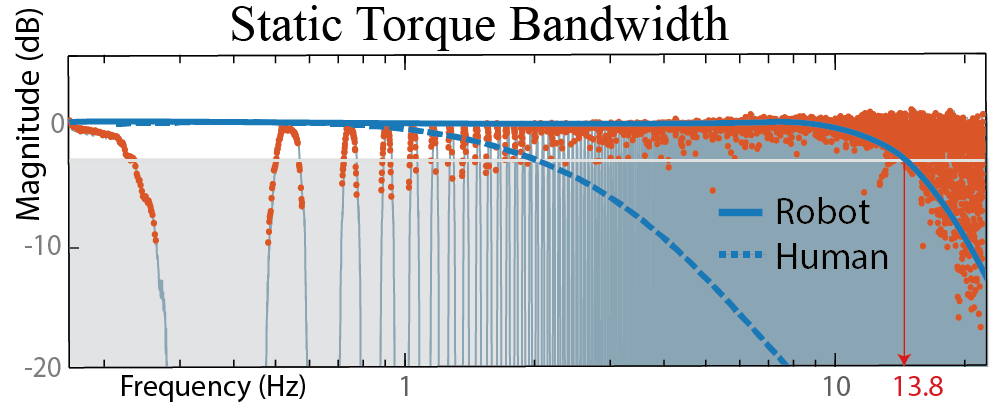}
\vspace{-9pt}
\caption{A static torque response to a requested chirp signal was plotted to determine the maximum effective control bandwidth of the \textit{Blue} system. The bandwidth was found to be 13.8 Hz which is greater than the bandwidth of human biceps muscle shown in dashed blue (2.3Hz \cite{aaron1976comparison}). Raw data is shown as faded, peaks are tracked in orange, robot torque response is in solid blue. Intersections between curves and the grey region represent frequencies beyond effective control (-dB). \bf{[video on website]}}
\label{fig:torqueBandwidth}
\vspace{-10pt}
\end{figure}

Torque bandwidth of a 2-DoF arm link was measured by grounding the actuator output to two 20 kg strain-based load cells whose signal is amplified by dual instrumentation amplifiers and then sampled by a 14bit ADC with digital low-pass filtering, passing the data in real time at 400 Hz to a central computer. A 10Nm chirp signal was commanded from 0.1 to 60 Hz over 300 seconds. As seen in Figure \ref{fig:torqueBandwidth}, non-linear harmonics in the output frequency response caused output torque to occasionally peak at unity-gain. A conservative estimate of bandwidth torque follows the roll-off of the sampled peaks, resulting in an estimated control bandwidth of 13.8 Hz. The human performance limit in an anthropomorphically analogous setting is 2.3 Hz \cite{aaron1976comparison}. 

\subsection{Thermal Experiments}
\begin{figure}[]
\centering
\vspace{8pt}
\includegraphics[width=1\linewidth]{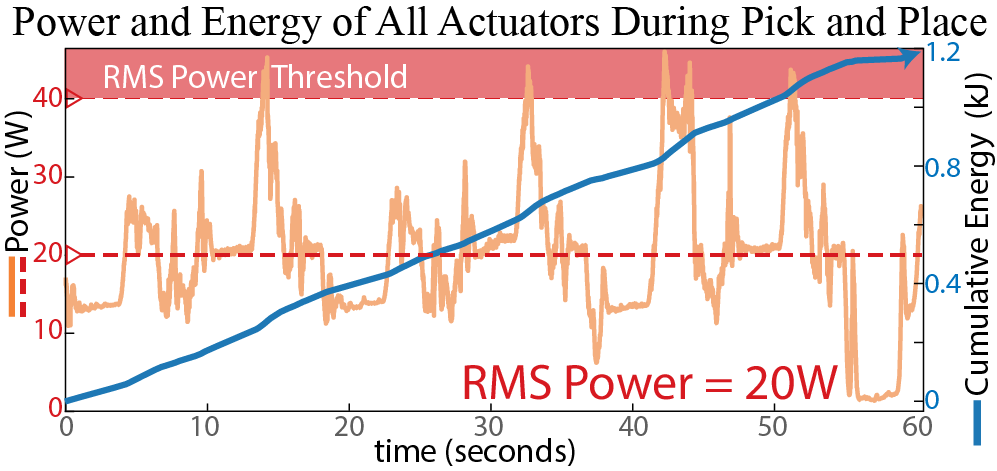}
\vspace{-9pt}
\caption{Inherent compliance and lower torque-density in \textit{Blue}'s actuators means that it is most effective when moving like we do. Humans rarely hold payloads at full extension and manipulate objects close to the body. In this figure, total power in all arm motors is tracked during a typical pick-and-place task. The robot can peak power output into the `thermal threshold' for short durations as long as RMS Power is kept below 40W}
\label{fig:heatPowerTotal}
\vspace{-10pt}
\end{figure}

\begin{figure}[]
\centering
\vspace{8pt}
\includegraphics[width=1\linewidth]{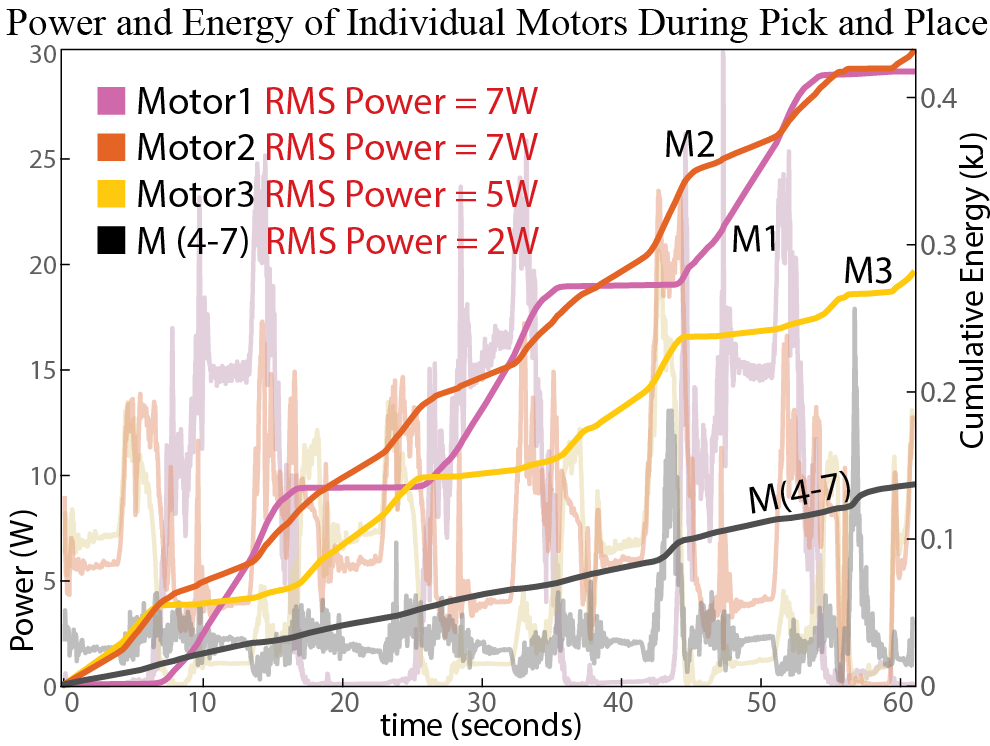}
\vspace{-9pt}
\caption{Heat generation strongly informs arm behavior and design for thermal dissipation. Energy is shown here in solid colors, instantaneous power is shown faded). During a pick-and-place task 90\% of total power is split between first three motors located in the Base (M1 - pink) and Shoulder (M2 - orange, M3 - yellow). Power used in remaining arm motors (M4, M5, M6, and M7) is summed in grey. Pick and place video on website}
\label{fig:heatPowerEach}
\vspace{-10pt}
\end{figure}

Maximizing backdrivability performance for QDD requires running motors near their thermal limits. Heat is generated (almost entirely) within the motors from $I^2R$ resistive losses. Achieving peak torque involves driving motors past their continuous thermal limits and motivates thermal testing. Average per-motor power consumption was evaluated then combined with measured thermal dissipation constants to inform peak capabilities of the robot.

Per-motor power was measured during a 60-second repetitive pick-and-place robot motion. Total RMS motor power is 20 Watts for `normal' movement with no payload (seen in Figure \ref{fig:heatPowerTotal}).

As shown in Figure \ref{fig:heatPowerEach}, 90\% of power is split between the proximal three arm motors (Base - M1; and Shoulder - M2, M3). Shoulder motors (M2, and M3) dissipate comparable amounts of heat and can be treated conservatively as a lumped thermal model to plan peak (`burst') arm capabilities in terms of \% duty cycle. Integrated motor power is shown in solid colors, suggesting that simple average power models can be used for normal movement. 

The base motor (M1) is bolted directly to a large aluminum base-plate which acts as a heat sink, providing significant thermal overhead. 
Within the body of the arm, thermal dissipation constants were measured at (0.93 W/\textdegree C with a fan, and 0.3 W/\textdegree C without a fan).
With the fan engaged, shoulder motors can dissipate 40W continuous at 70\textdegree C, resulting in a combined $\approx$ 20Nm continuous output torque (0.5kg payload fully outstretched), or 20 Watts of cooling overhead if the established average from Figure \ref{fig:heatPowerTotal} is respected.

Assuming a 20 Watt RMS `resting' power, one can temporarily dump heat into the shoulder motors (estimated 1103 Ws/\textdegree K heat capacity), producing an excess of 35Nm of torque for 23\% of the time (duty cycle): capable of holding a 2kg payload at full extension with a 10 Nm dynamic overhead for upwards of 2 minutes before seeing a $>10$\textdegree C rise in motor temperature, and having to `rest' for 7 minutes. Shorter bursts are possible with less time spent resting, as long as total average power doesn't exceed 40 Watts.

\subsection{Teleoperation Tasks}
A set of human tasks were attempted under teleoperation to evaluate the robot's qualitative performance, one of which is shown in Figure \ref{fig:vrTeleop}. Challenges included gauging object depth through the constrained camera feed of the robot and commanding interaction forces, since a single controller tune was used and the only input is a 6-DOF position target in task-space. Successful tasks included operating an espresso machine, picking up m5 bolts, cleaning a table with paper towels, and decluttering.

\section{Manufacturability and Cost Analysis}
\label{sec:manufacturing}

%Cost and manufacturability of the \textit{Blue} robot arm were tracked at every step of the design process. 
We designed all plastic components to be injection molded for high volumes, metal pieces to be planar and simple to machine, and hardware to be commodity or sourced from existing high volume products such as bicycles, 3D printers, and drones. 
In the lab we fabricated seven \textit{Blue} arms (of the version presented in this paper) for testing. Arms can be deployed minimally as single units with base flat on table.

\subsection{Prototype Cost}
%There are a total of $492$ components in each arm.
During a $7$ unit fabrication run in-house, Bill of Materials (BoM) cost for each arm was tracked at \$$3328$ as shown in Table \ref{prototypeCost}. 
Plastics were printed on \textit{Markforged Onyx One} and \textit{Monoprice MP Select Mini} 3D printers.
About 75\% of our prototype costs were motors and driver boards.
%Bearings were sourced directly from Chinese and Taiwanese manufacturers. 
Machined components were sourced from a machine shops globally. 
Assembly took about $6$ hours per robot arm. 
Assembly costs were not factored into Table \ref{prototypeCost}.

\begin{table}[h]
\caption{Motor Specs \& Prototyping cost for \textit{Blue} arms}
\label{prototypeCost}
\vspace{-9pt}
\centering
\includegraphics[width=1\linewidth]{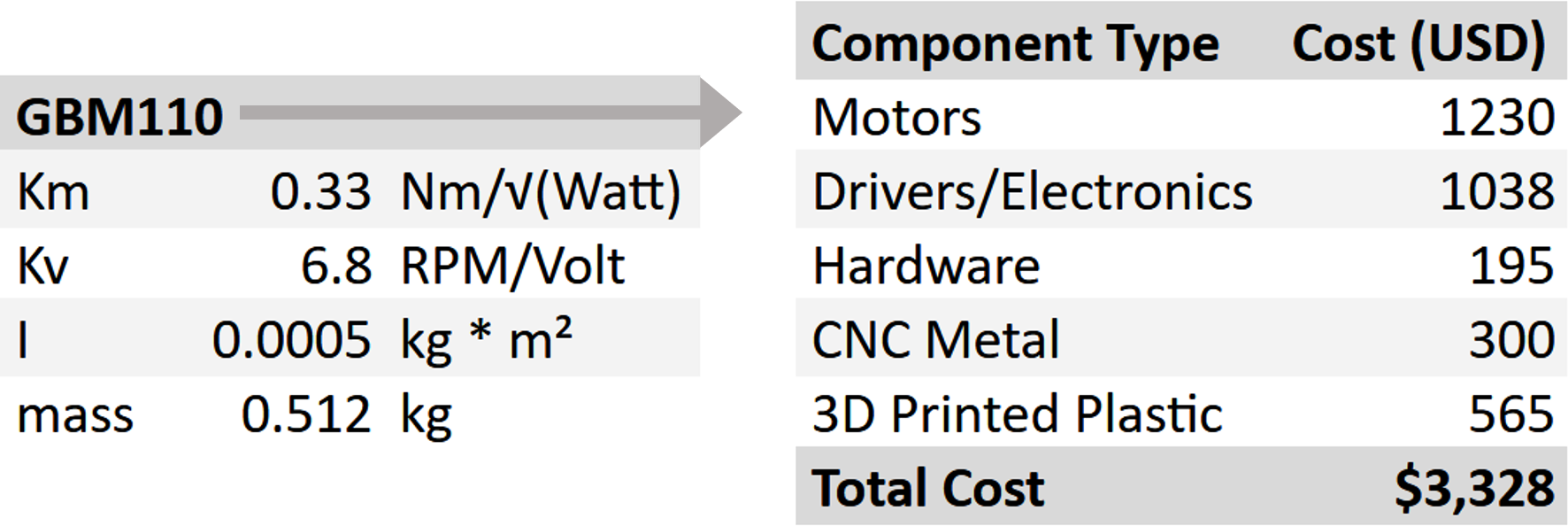}
\end{table}

\subsection{Manufacturing and Scaling Cost Analysis}
To fully evaluate a low-cost design paradigm for manipulation, we worked with three Contract Manufacturing (CM) companies to identify the costs of \textit{Blue} arms produced at scale. We received quotes from three CM's in the California Bay Area from which we based the estimates presented in Table \ref{table:manu}. 
Creating tooling for the $25$ unique plastic components is estimated at \$$160,000$. Other CM bring-up costs and Non Recoverable Engineering expenses (NRE's) could total \$$12,000$. 
As shown in Table \ref{table:manu},
the end cost to consumers (assuming additional operational margins) can be kept within our \$$5000$ goal range if producing at volumes above $1500$ \textit{Blue} arms.

%\vspace{9pt}

\begin{table}[]
\caption{Manufacturing BoM cost breakdown for \textit{Blue} arms}
\label{table:manu}
\vspace{-9pt}
\centering
\includegraphics[width=1\linewidth]{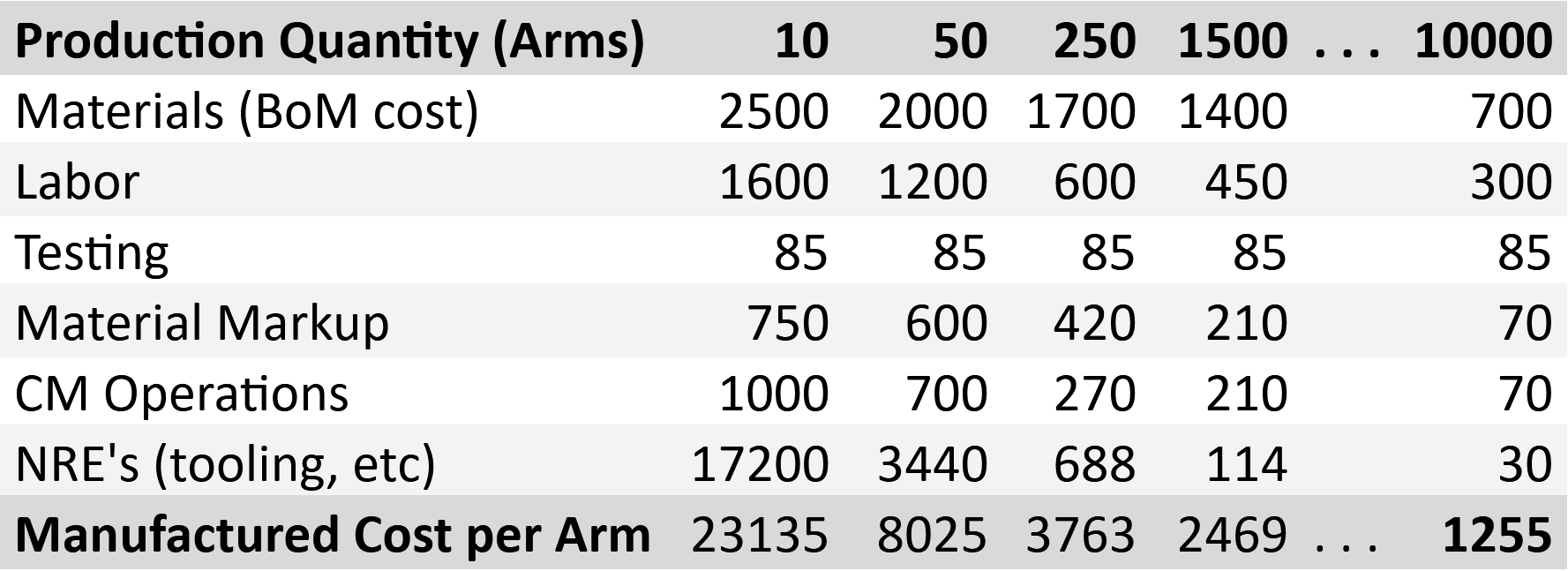}
\end{table}

%\vspace{-12pt}

\section{Discussion and Future Work}
\label{sec:futureWork}
% The \textit{Blue} system would benefit from further development on mechanical, firmware, and control.
Future mechanical work includes increasing robustness by reducing timing belt skips, implementing a fiber reinforced thermoset polymer shell to avoid long-term structural creep, and
adding slip rings to enable continuous rotation,
eliminate hard stops, and reduce cable fatigue.
Software improvement includes
disturbance observers using onboard accelerometers, and
fully automatic startup calibration procedures.
%visual servoing to implement high accuracy closed-loop precision,
%and more intuitive inverse kinematics for better VR teleoperation.

Benchmarking compliance across robot platforms would benefit the manipulation community.
Future work includes testing and comparing force control capabilities of commercially available robots by measuring Z-width (which represents achievable impedance across frequencies) \cite{weir2008measuring}.

%%%%%%%%%%%%%%%%%%%%%%%%%%%%%%%%%%%%%%%%%%%%%%%%%%%%%%%%%%%%%%%%%%%%%%%%%%%%%%%%
%\section*{APPENDIX}

\vspace{-4pt}
\section*{Acknowledgments}
\vspace{-4pt}
The Robot Learning Lab (RLL) is part of the University of California, Berkeley.
This work is supported in part by the National Science Foundation Graduate Research Fellowship under Grant No. DGE 1752814, NSF Career Grant No. 1351028, the Toyota Research Institute, and by the Bakar Fellows Program.
The authors would like to recognize the help of the UC Berkeley Student Machine Shop, Michael McKinley, Morgan Quigley, Roshena Macpherson, Justin Yim, Jeffrey Mahler, and Dominick Kofi Yaate Quaye.
\vspace{-9pt}
%%%%%%%%%%%%%%%%%%%%%%%%%%%%%%%%%%%%%%%%%%%%%%%%%%%%%%%%%%%%%%%%%%%%%%%%%%%%%%%%
%\bibliographystyle{IEEEtranS}%alphabetical refs
\bibliographystyle{IEEEtran}%non-alphabetical
\bibliography{arm_library}

\end{document}